\documentclass{llncs}
\usepackage{makeidx}  
\usepackage{graphicx}
\usepackage{siunitx}
\usepackage{amsmath,amsfonts,amssymb}
\usepackage{caption}
\usepackage{subcaption}
\usepackage{color}

\begin{document}
\pagestyle{headings}  
\title{GP-Unet: Lesion Detection from Weak Labels with a 3D Regression Network}

\author{Florian Dubost\inst{1,2,3} \and Gerda Bortsova\inst{1,2,3} \and Hieab Adams\inst{3,4} \and Arfan Ikram\inst{3,4,5} \and Wiro Niessen\inst{1,2,3,6} \and Meike Vernooij\inst{3,4} \and Marleen De Bruijne\inst{1,2,3,7}}

%
\institute{Biomedical Imaging Group Rotterdam, Erasmus MC, Rotterdam, The Netherlands
\and Department of Medical Informatics, Erasmus MC, Rotterdam, The Netherlands
\and Department of Radiology, Erasmus MC, Rotterdam, The Netherlands
\and Department of Epidemiology, Erasmus MC, Rotterdam, The Netherlands 
\and Department of Neurology, Erasmus MC, Rotterdam, The Netherlands
\and Imaging Physics, Faculty of Applied Sciences, TU Delft, The Netherlands
\and Department of Computer Science, University of Copenhagen, Denmark}

\maketitle

\begin{abstract}
We propose a novel convolutional neural network for lesion detection from weak labels. Only a single, global label per image - the lesion count - is needed for training. We train a regression network with a fully convolutional architecture combined with a global pooling layer to aggregate the 3D output into a scalar indicating the lesion count. When testing on unseen images, we first run the network to estimate the number of lesions. Then we remove the global pooling layer to compute localization maps of the size of the input image. We evaluate the proposed network on the detection of enlarged perivascular spaces in the basal ganglia in MRI. Our method achieves a sensitivity of $62\%$ with on average $1.5$ false positives per image. Compared with four other approaches based on intensity thresholding, saliency and class maps, our method has a $20\%$ higher sensitivity.
\end{abstract}

\section{Introduction}
This paper addresses the problem of the detection of small structures in 3D images. We aim at developing a machine learning method requiring the least possible annotations for training. Several deep learning techniques \cite{3dunet,voxresnet,calci} have recently been proposed for 3D segmentation. These methods use fully convolutional networks (FCN) \cite{shel} with a downsampling and upsampling path and can therefore detect small regions and fine details. Although efforts have been made to reduce the amount of annotations required with e.g, sparse annotations \cite{3dunet}, these techniques still need pixel-wise annotations for training. Acquiring those is very time-consuming and often not feasible for large datasets.
\let\thefootnote\relax\footnote{The final publication is available at Springer via http:://dx.doi.org/10.1007/978-3-319-66179-7 25}

In \cite{actMap} the authors propose 2D localization networks requiring only global image labels for their training. They aggregate the last features maps of their network into scalars thanks to a global pooling layer and can therefore compute a classification. Heatmaps can be obtained as a weighted sum of these feature maps. These heatmaps indicate which regions of the image contributed the most to the classification results. 
Because of the downsampling introduced by the pooling operations in the network, the heatmaps are several times smaller than the original input image, which makes it impossible to detect very small structures.
In \cite{hwang} a similar technique is applied to 3D CT lung data. This network splits into two branches, one with fully connected layers for classification, and the other with a global pooling layer for localization. The loss is computed as a weighted sum of these two terms. However, as in \cite{actMap}, this network is not suitable for the detection of small structures.

In this paper we propose a method to learn fine pixelwise detection from image-level labels. By combining global pooling with a fully convolutional architecture including upsampling steps as in the popular U-Net \cite{unet}, we compute a heatmap of the size of the input. This heatmap reveals the presence of the targeted structures. During the training, unlike \cite{actMap} or \cite{hwang} where the authors use a classification network, the weights of the network are optimized to solve a regression task, where the objective is to predict the number of lesions.

We evaluate our method on the detection of enlarged perivascular spaces (EPVS) in the basal ganglia in MRI. EPVS is an emerging biomarker for cerebral small vessel disease. The detection of EPVS is a challenging task - even for a human observer - because of the very small size and variable shape. 
In \cite{ramirez,park,gonzales1} the authors propose different EPVS segmentation methods. However detection of individual EPVS has not been addressed much in the literature. Only \cite{park} proposes an automatic method for this, but this method requires pixel-wise annotated scans for training.

\section{Methods}
\label{sec:methods}

Our method takes as input a full 3D MRI brain scan, computes a smaller smooth region of interest (ROI) and inputs it to a FCN which computes a heatmap revealing the presence of the targeted brain lesions. The FCN is trained with weak labels. Its architecture changes between training and testing but the optimized weights stay the same.

\subsection{Preprocessing}
\label{sec:preProcess}
Scans are first registered to MNI space. A binary mask segmenting a region of interest (ROI) in the brain is computed with standard algorithms \cite{deskian}. This mask is dilated and its borders are smoothed with a Gaussian kernel. After applying the mask, we crop the image around the ROI, trim its highest values and rescale its intensities such that the input to the network is $S \in [0,1]^{h \times w \times d}$.
\subsection{3D Regression Fully Convolutional Network}
\label{sec:regNet}

\begin{figure}[t]
\centering
\includegraphics[height=6cm]{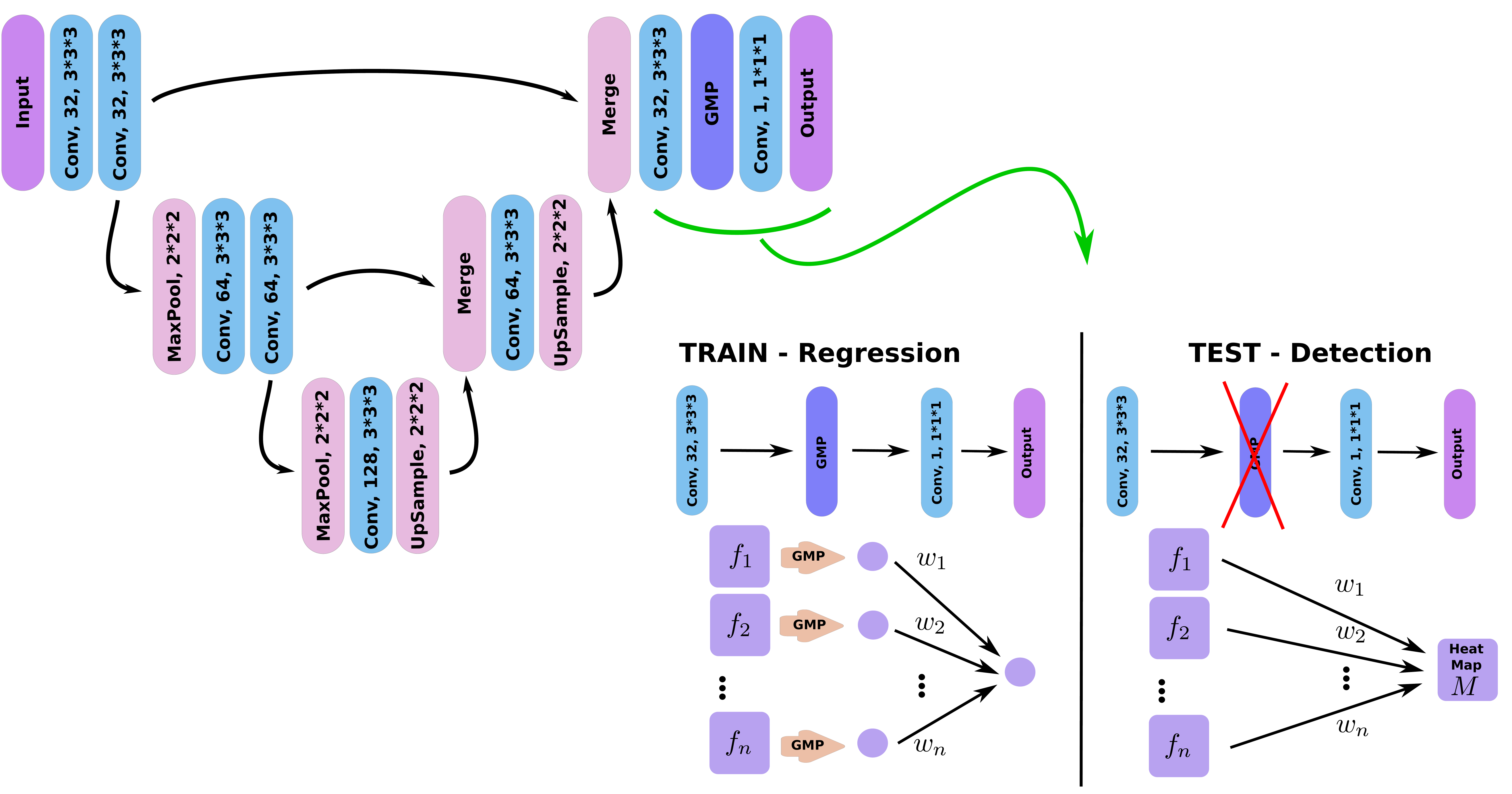}
\caption{\textbf{3D Regression FCN Architecture.} Top-left: network architecture (see Sec. \ref{sec:regNet}). Bottom-right: during training, the network is built to solve a regression problem and outputs $\hat{y} \in \mathbb{R}$ computed according to Eq.(\ref{eq:conv}). During testing, the global pooling layer is removed. Using Eq.(\ref{eq:classMap}), the network computes a heatmap $M \in \mathbb{R}^{h \times w \times d}$ of the same size as the network input volume $S$.}
\label{fig:arch}
\end{figure}
Fig. \ref{fig:arch} shows the architecture of our network. It is similar to the one of 3D U-Net \cite{3dunet} but is less deep and has, during training, a global pooling layer \cite{NIN} before the last layer. The use of this layer is detailed in the next section. Our network has $3 \times 3 \times 3$ convolutional layers. In the encoding path, these layers are alternated with $2 \times 2 \times 2$ pooling layers to downsample the image and increase the size of the receptive field. In the decoding path we use $2 \times 2 \times 2$ upsampling layers. We use padding. Skip connections connect the encoding and decoding paths. After each convolutional layer - except for the last one - we compute a rectified linear unit activation . The depth and number of feature maps of the network are set to fit the available GPU memory. See Fig. \ref{fig:arch} for the architecture details.

We change the last layers of our network between training and testing. The testing step is performed with a standard fully convolutional architecture. The output is a heatmap of the size of the input. During training, we use only global image labels. To compute the loss function we need therefore to collapse the image output into a scalar. For this purpose, during training only, we introduce a global pooling layer before the last layer. In others word, we train a regression network and transform it to a segmentation network for testing (See Fig.\ref{fig:arch}). We detail this below.

\subsubsection{Training - Regression Network.}

After the last convolutional layer, instead of stacking the voxels of each feature map and feeding them to a fully connected layer, we use a global pooling layer, which computes one pooling operation for each feature map. The resulting output of such a layer is a vector of scalars $x \in \mathbb{R}^{n}$, with $n \in \mathbb{N}$ the number of features maps of the preceding layer.

Let $f_{i} \in \mathbb{R}^{h \times w \times d}$, for $i \in \{1,2,..,n\}$, be the i-th feature map received by the global pooling layer.
We can write the global pooling operation as $g(f_{i}) = x_{i}$, with $x_{i} \in \mathbb{R}$.
Let $G$ be the underlying mapping of the global pooling layer and $f \in \mathbb{R}^{n \times h \times w \times d}$ the vector containing the $n$ feature maps $f_{i}$. We have $G(f) \in \mathbb{R}^{n}$ and in the case of global max pooling we can write $(G(f))_{i} = \max\limits_{x,y,z} f_{i,x,y,z}$.

A convolution layer with a single output feature map $\hat{y}$ and a filter size of $1 \times 1 \times 1$ can be written as 
\begin{equation}
\label{eq:conv}
\hat{y} = \sum_{i} w_{i} h_{i} + b,
\end{equation}
with $w_{i} \in \mathbb{R}$, the weights, $b$ the bias and $h_{i}$ the i-th feature map of the preceding layer. Considering the layer following the global pooling layer, we can write Eq. (\ref{eq:conv}) as
\begin{equation}
\label{eq:output}
\hat{y} = \sum_{i} w_{i} (G(f))_{i} + b = w.G(f) + b,
\end{equation}
with $.$ denoting the scalar product, $\hat{y} \in \mathbb{R}$, $b \in \mathbb{R}$ and $w \in \mathbb{R}^{n}$. The output of the network $\hat{y}$ is the estimated lesion count. There is no activation function applied after this last convolutional layer. To optimize the weights, we compute a mean squared loss $l = \frac{1}{m}\sum_{t=1}^{m}(\hat{y}_{t} - y_{t})^2$, where $\hat{y}_{t} \in \mathbb{R}$ is the predicted count, $y_{t} \in \mathbb{N}$ the ground truth for volume $S_{t}$ and $m$ the number of samples.

\subsubsection{Testing - Detection Network.}
During testing we remove the global pooling layer, but we do not change the weights of the network. The $h_{i}$ of Eq. (\ref{eq:conv}) are the feature maps $f_{i}$ defined earlier and the bias $b^{l}$ can be omitted. We can thus rewrite Eq. (\ref{eq:output}) as
\begin{equation}
\label{eq:classMap}
M = \sum_{i} w_{i} f_{i},
\end{equation}
with $M \in \mathbb{R}^{h \times w \times d}$ the new output of the network. $M$ is a heatmap indicating the location of the targeted lesions. Note that computation of $M$ is mathematically equivalent to the class map computation proposed in \cite{actMap}. However $M$ has the same size as the input $S$ and the weights $w_{i}$ are optimized for a regression problem instead of classification. 

After computing the heatmap, we need to select an appropriate threshold before we can compute a segmentation and retrieve the location of the targeted lesions. This efficiently be done by using the estimate of the lesion count $\hat{y}$ provided by the neural network: we only need to keep the global pooling layer while testing. The threshold is then selected so that the number of connected components in the thresholded heatmap is equal to the lesion count.

\section{Experiments}
We evaluate our method on a testing set of 30 MRI scans and compute the sensitivity, false discovery rate and average number of false positive per image.
\subsubsection{Data}
Our scans were acquired with a 1.5 Tesla scanner. Our dataset is a subset of the Rotterdam Scan Study \cite{RSS} and contains 1642 3D PD-weighted MRI scans. Note that in our dataset PD-weighted images have a signal intensity similar to T2-weighted images, the modality commonly used to detect EPVS. The voxel resolution is $0.49 \times 0.49 \times 0.8 \text{mm}^{3}$. Each of the scans of dataset is annotated with a single, global image label: a trained observer counted the number of EPVS in the slice of the basal ganglia showing the anterior commissure. In a random subset of $30$ scans, they marked the center of all EPVS visible in this same slice. These $30$ scans are kept for testing set. The remaining dataset is randomly split into the following subsets: $1289$ scans for training and $323$ for validation. 
\begin{figure}[t]
\centering
\includegraphics[height=3cm]{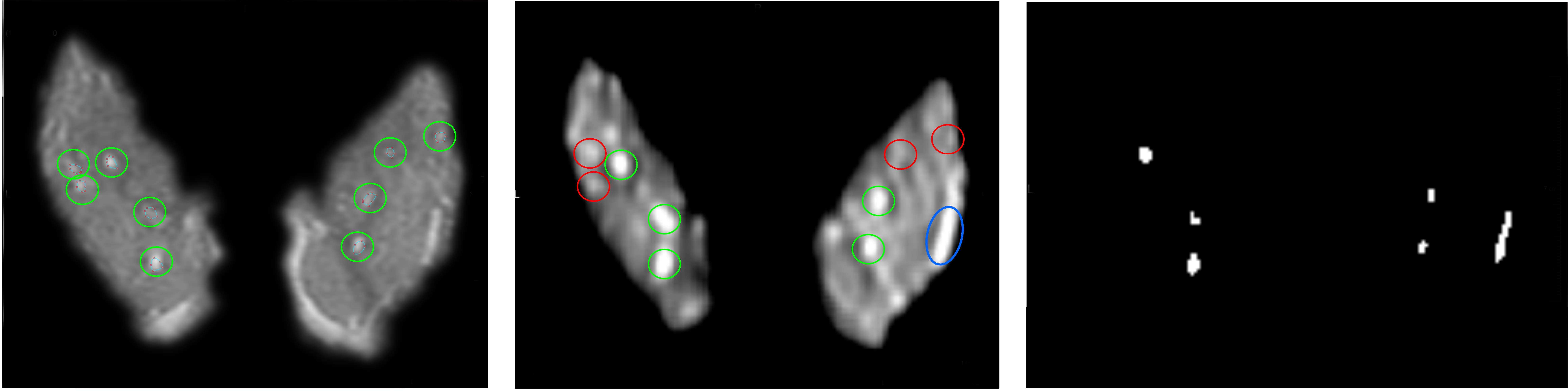}
\caption{\textbf{Examples of EPVS detection in the basal ganglia.} Left: Ground truth on the intensities - the lesions are circled in green. Center: Heatmap from the neural network. True positives in green, false positives in blue and false negatives in red. Right: Segmentation of the heatmap.}
\label{fig:data}
\end{figure}  
\subsubsection{Experimental Settings}
\label{sec:param}
The initial ROI segmentation of the BG is computed with the subcortical segmentation of FreeSurfer \cite{deskian}.
For registration to MNI space we use the rigid registration implemented in Elastix \cite{elastix1} with the mutual information as similarity measure and default settings. The Gaussian blurring kernel has a standard deviation $\sigma = 2$.
The preprocessed image $S$ has a size of $168 \times 128 \times 84$ voxels.
We trim its $1\%$ highest values.
We initialize the weights of the CNN by sampling from a Gaussian distribution, use Adadelta \cite{adadelta} for optimization and augment the training data with randomly transformed samples.
A lesion is counted as detected if the center of a connected component in the thresholded heatmap is located within a radius of $x$ pixels of the annotation. In our experiments we set $x=3$ which corresponds to the average radius of the largest lesions.
We implemented our algorithms in Python with Keras and Theano libraries and ran the experiments on a Nvidia GeForce GTX 1070 GPU.
\begin{table}[h]
\setlength{\tabcolsep}{10pt}
\caption{\textbf{Results of our method in comparison to the baselines.} TPR stands for true positive rate (i.e. sensitivity), FPav is the number of false positives per image in average and FDR stands for false discovery rate. The comparing methods are described in Sec. \ref{sec:base}.}
\begin{center}
\begin{tabular}{c c c c}
\hline\rule{0pt}{12pt}
Method & TPR & FPav & FDR\\
\hline\rule{0pt}{12pt}
Intensities	 (a)				&	40.6	&	2.3	&	59.8 \\
Saliency (b)					&	39.8	&	2.7	&	54.2 \\
Saliency FCN (c)			&	18.7	&	3.3	&	70.2 \\
Regression (d)             &  19.6			&	3.2	&	70.5	\\
Regression FCN	(e)		&	54.8	&	1.9		&	37.7 \\
Intensities + Reg FCN (f)	&	\textbf{62.0}	&	\textbf{1.5}	&	\textbf{31.4} \\
\hline\rule{0pt}{12pt}
\end{tabular}
\end{center}
\label{table:results}
\end{table}
\subsubsection{Baseline Methods}
\label{sec:base}
We compare our method with four conventional approaches, using the same preprocessing (Sec. \ref{sec:preProcess}). 
The first method (a) thresholds the input image based on it intensities: EPVS are among the most hyperintense structures in the basal ganlgia.
Both second and third methods compute saliency maps \cite{saliency} using a neural network. Saliency maps are obtained by computing the gradient of the regression score with respect to the input image. These maps can be easily computed given any classification or regression network architecture. The third method (c) uses the same network architecture as ours, the second one (b) uses a regression network without the upsampling path. Finally we compared with a method similar to \cite{actMap} where we replace the original classification objective with a regression. This network is the same as ours but without the upsampling part.
The thresholds are chosen based on the lesion count estimated by each network as explain in  Sec. \ref{sec:methods}.
\subsubsection{Results}
\label{sec:results}
\begin{figure}[t]
\centering
\includegraphics[height=2.5cm]{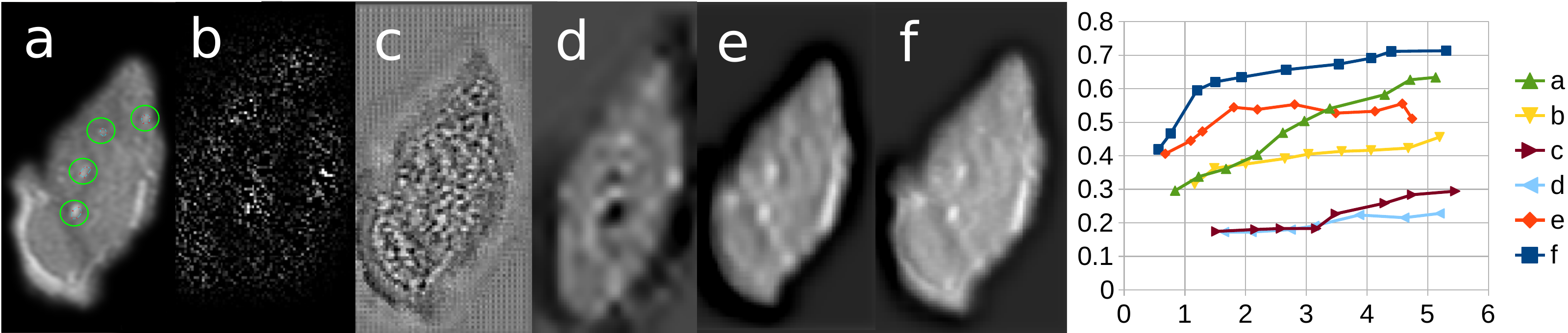}
\caption{\textbf{Comparison with baselines.} Left: Heatmaps of the different methods. EPVS are circled in green on the original image (a). The saliency maps (b, c(FCN)) are very noisy. Our method (e) has finer details than a network without upsampling path (d) similar to \cite{actMap}. Combining the intensities with our heatmap (f) offers the best results. Right: Free-ROC curves for each method, showing the FPav on the x-axis and the sensitivity on the y-axis.}
\label{fig:comp}
\end{figure}

In Table \ref{table:results} we report sensitivity, false discovery rate and average false positive per image (FPav). Our method (e) detects $54.8\%$ of the lesions with on average $1.9$ false positive per image. We found that the performance could be improved further by thresholding the weighted sum of the heatmap and the original voxel intensities of the image (f). This improves the sensitivity to $62.0\%$ with $1.5$ FPav. Considering the sensitivity, our method outperforms the other baseline methods by more than $20\%$  (Table. \ref{table:results}). 
By modifying the heatmap threshold to overestimate the lesion count, we can increase the sensitivity further - $71.1\%$ - at the price of more false positive - $4.4$ in average and a false discovery rate of $60\%$ (Fig. \ref{fig:comp}). 
\section{Discussion}
Our experiments show that using a fully convolutional network and optimizing its weights for a regression problem gives the best results on our dataset (Fig. \ref{fig:comp} and Table \ref{table:results}). 

The methods in \cite{hwang,actMap} produce coarse localization heatmaps on which it is impossible to locate small structures. In our early experiments we computed similar networks, optimized for classification. The resulting heatmaps highlight the whole ROI, which does not help to solve the detection problem. Using a regression network, the more precise output allows us to extract finer-scale information, but without an upsampling path, the results are still imprecise (see Fig. \ref{fig:comp} method (d)). 
In our experiments, saliency maps \cite{saliency} perform similarly to voxel intensity. By construction these maps are noisy (Fig. \ref{fig:comp} method (b)). 

Considering other works in automatic detection of EPVS, in \cite{park} the authors compute an EPVS segmentation using a random forest classifier. They train this classifier with pixelwise ground truth annotations. They also use a 7T scanner which produces high resolution images and eases small lesion detection. They report a sensitivity of $78\%$ and a false discovery rate of $55\%$. This is better than our $71\%$ sensitivity and $60\%$ false discovery rate. However contrary to \cite{park} which requires pixelwise annotation, our method uses only weak labels for training. In \cite{gonzales1,ramirez}, the authors also compute segmentations, but the evaluation is limited to a correlation with a five-category visual score.

Overall we consider that our detection results are satisfactory, considering that, during training, our method did not use any ground truth information on the location or shape of the lesions, but only their count within the whole ROI.
\section{Conclusion}
We presented a novel 3D neural network to detect small structures from global image-level labels. Our method combines a fully convolutional architecture with global pooling. During training the weights are optimized to solve a regression problem, while during testing the network computes lesion heatmaps of the size of the input. We demonstrated the potential of our method to detect enlarged perivascular spaces in brain MRI. We obtained a sensitivity of $62\%$ and on average $1.5$ false positives per scan. Our method outperforms four other approaches with an increase of more than $20\%$ in sensitivity. We believe this method can be applied to the detection of other brain lesions.

\subsubsection{Acknowledgments}
This research was funded by  The Netherlands Organisation for Health Research and Development (ZonMw) Project 104003005.
%
%
%

\end{document}